\documentclass[letterpaper, 10 pt, conference]{ieeeconf}  

\IEEEoverridecommandlockouts                              

\usepackage{graphics} 
\usepackage{epsfig} 
\usepackage{times} 
\usepackage{amsmath} 
\usepackage{amssymb}  
\usepackage{tabularx}
\usepackage{verbatim}

\usepackage[%
  vlined,
  ruled
]{algorithm2e}

\title{\LARGE \bf
Deep Adversarial Reinforcement Learning for Object Disentangling
}

\author{Melvin Laux$^{1}$, Oleg Arenz$^{1}$, Jan Peters$^{1,2}$ and Joni Pajarinen$^{1,3}$
\thanks{$^{1}$Intelligent Autonomous Systems, TU Darmstadt, Germany}%
\thanks{$^{2}$MPI for Intelligent Systems, Tuebingen, Germany}%
\thanks{$^{3}$Learning for Intelligent Autonomous Robots, Tampere University}%
}

\begin{document}

\maketitle
\thispagestyle{empty}
\pagestyle{empty}

\begin{abstract}

Deep learning in combination with improved training techniques and
high computational power has led to recent advances in the field of
reinforcement learning (RL) and to successful robotic RL applications
such as in-hand manipulation.
However, most robotic RL relies on a
well known initial state distribution. In real-world tasks, this 
information is however often not available. For example, when
disentangling waste objects the actual position
of the robot w.r.t. the objects may not match the positions
the RL policy was trained for.
To solve this problem, we present a novel adversarial reinforcement learning
(ARL) framework. The ARL framework utilizes an adversary,
which is trained to steer the original agent, the protagonist, to
challenging states. We train the protagonist and the adversary jointly 
to allow them to adapt to the changing policy of their opponent.
We show that our method can generalize from
training to test scenarios by training an end-to-end system for robot
control to solve a challenging object disentangling task. 
Experiments with a KUKA LBR+ 7-DOF robot arm show that our approach
outperforms the baseline method in disentangling when starting from
different initial states than provided during training.

\end{abstract}

\section{INTRODUCTION}
Deep reinforcement learning (DRL) methods have achieved remarkable performance in playing Atari games \cite{mnih2013playing, mnih2015humanlevel}, Go \cite{silver2016mastering}, Chess~\cite{AlphaZero}, Dota 2~\cite{Dota2} or Starcraft 2~\cite{Starcraft2}. However, tackling such tasks required a large number of policy roll-outs---typically in the order of billions. This huge sample complexity is in stark contrast to the number of roll-outs that are feasible when training a policy on a real robot, which is typically in the range of tens or hundreds.
Still, deep reinforcement learning has proved its usefulness also for robot applications such as in-hand manipulation~\cite{akkaya2019}, object manipulation~\cite{zhu2018}, quadruped locomotion~\cite{haarnoja2018soft2} or autonomous driving~\cite{bojarski2016end}. These results were made possible by using a variety of techniques to reduce the required number of real-world interactions, such as (pre-)training in simulation, initializing from demonstrations, or incorporating prior knowledge in the design of the Markov Decision Process (MDP). However, policies that are trained with few real-world interactions can be prone to failure when presented with unseen states. It is, thus, important to carefully set the initial states of the robot and its environment so that it also includes challenging situations.

\begin{figure}
  \centering
    \includegraphics[width=0.45\textwidth]{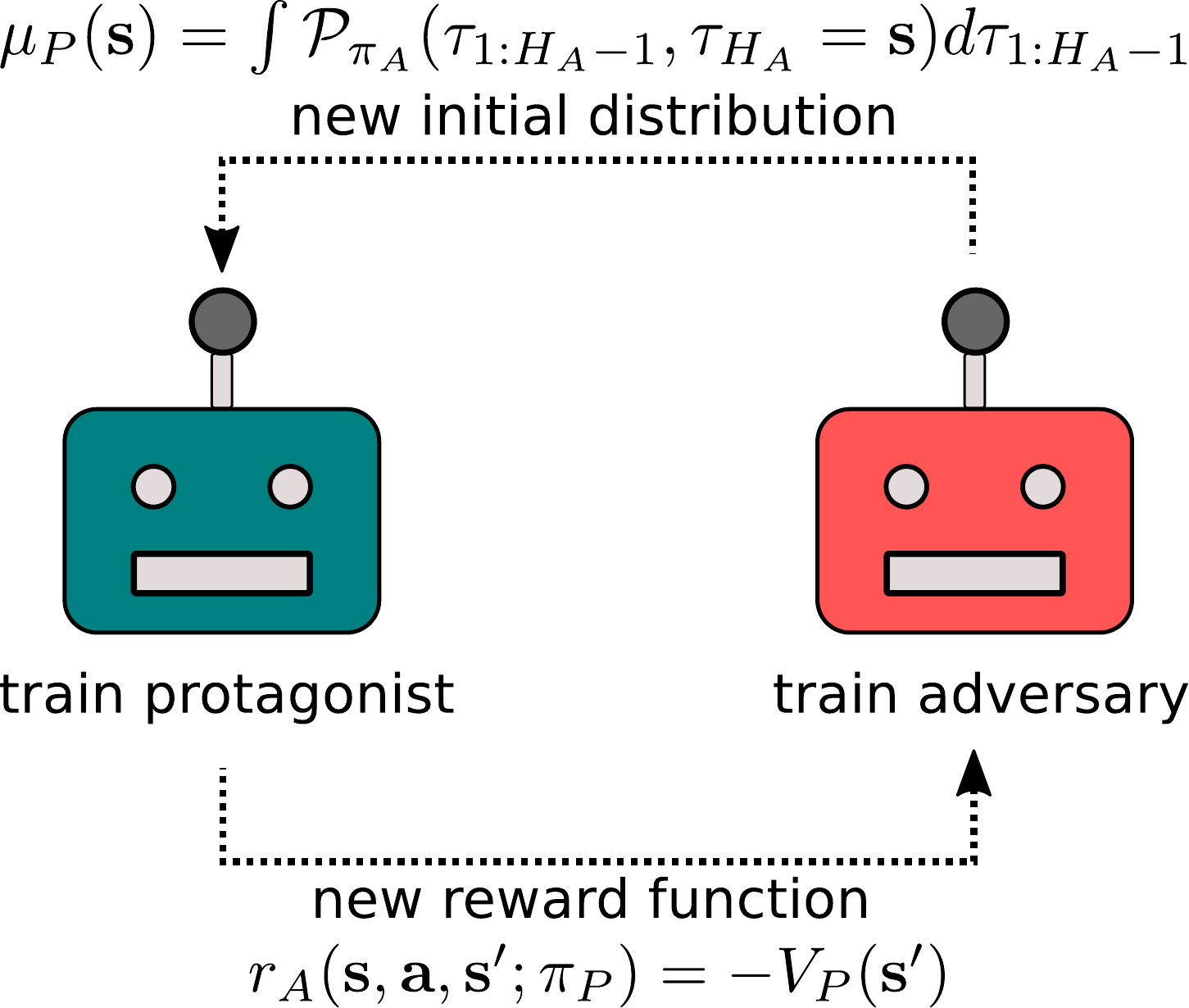}
  \caption[A high-level illustration of the ARL learning process]{During training our Adversarial Reinforcement Learning (ARL) approach loops over two steps. First, ARL trains the adversary to maximize its current reward function, which depends on the protagonist's current value function $V_P(\mathbf{s}')$. The adversary's updated policy generates a new challenging initial state distribution $\mu_P(\mathbf{s})$ for the protagonist. Next, the protagonist is trained to maximize expected reward when starting from this new state distribution. The ARL process leads to a protagonist that can cope with a challenging initial state while keeping the initial state distribution valid: the adversary was able to explore those states.}
  \label{fig:arl_vis}
\end{figure}

\begin{figure}
      \centering
      \label{fig:iiwa_upright}
      \includegraphics[width=0.49\linewidth]{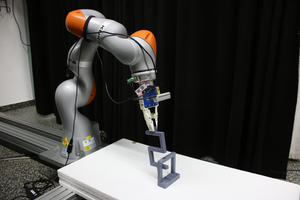}
      \centering
      \label{fig:shapes}
      \includegraphics[width=0.49\linewidth]{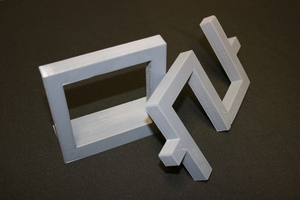}
      \caption[Experimental hardware setup]{We show the benefits of
        our ARL approach in both simulation and in a real robot task
        where the robot tries to disentangle an object from another
        one. Left: The KUKA LBR+ 7-DOF robot arm with attached SAKE
        gripper disentangling an "S-shape" object from an "O-shape"
        object. Right: The 3D printed "O-shape" and "S-shape"
        objects.}
    \label{fig:hardware_setup}
    
\end{figure}

Furthermore, unlike simulations, the state of robot and environment can not be easily reset. Using a human to manually set up the environment after each roll-out can often introduce a significant bottleneck in the data collection. Hence, successful applications of reinforcement learning to robotics typically involved tasks that can be easily reset to their initial states, for example playing table tennis against a ball-cannon~\cite{Kober2012} or grasping objects from a bin~\cite{levine2016learning}. Based on these observations, we investigate how to \emph{automatically} set up the environment to a \emph{challenging} state for the agent.

We present a novel adversarial learning framework for reinforcement learning algorithms: Adversarial reinforcement learning (ARL). This framework helps to learn RL policies that can generalize to new situations by employing an adversarial policy to steer the agent to unknown or difficult situations. Through this adversary driven exploration of the state space, the agent is more likely to learn a more general policy than by using purely random exploration. 
While we assume irreducibility of the MDP to prevent the adversary from creating impossible tasks for the protagonist, we do not make any assumptions on the reinforcement learning algorithm.
An overview of the framework is illustrated in Figure~\ref{fig:arl_vis}.

We investigate the task of disentangling waste~\cite{pajarinen2020probabilistic}. There is an enormous potential of robots in the waste processing industry by reducing human labor and allowing for better segregation and thus recycling of waste. Furthermore, there is high interest in the nuclear industry to employ robots for segregating nuclear waste, which is inherently dangerous for humans. To apply reinforcement learning to waste disentangling it would be desirable to automate the generation of challenging entanglements. However, designing such procedure by hand is not straightforward and, thus, it is interesting to investigate our learning-based approach.

\textbf{Contribution.}
\begin{itemize}
	\item We propose an adversarial framework (ARL) for reinforcement learning that alternately trains an agent for solving the given task and an adversary for steering the protagonist to challenging states.
 \item We evaluate ARL in simulation as well as on a real robot on the object disentangling task. 
For the robot experiment, a KUKA iiwa R820 in combination with a SAKE EzGripper is tasked to disentangle two 3D-printed objects as shown in Figure~\ref{fig:hardware_setup}. 
\end{itemize}

\section{RELATED WORK}
\label{sec:related}
\textbf{Adversarial learning.}
The idea of embedding adversaries into the training process has been shown to work successfully in the field of supervised learning \cite{goodfellow2014generative, goodfellow2014explaining} and imitation learning \cite{ho2016generative}. Adversarial approaches have also been previously incorporated in RL methods to learn more robust policies by adding adversary-generated noise to the policy's actions during training \cite{pinto2017robust, pan2019risk}, using an adversarial robot to interfere with the protagonist \cite{pinto2017supervision} or training an adversary to perturb the protagonist's observations to cause the agent to fail \cite{pattanaik2017robust, huang2017adversarial, mandlekar2017adversially}. Gleave et al. investigate the vulnerability of DRL methods against adversarial attacks in the context of multi-agent environments \cite{gleave2019adversarial}. They show how attacks can be carried out effectively by training adversarial policies to generate natural adversarial observation via self-play. \textit{Robust adversarial reinforcement learning} (RARL) framework by Pinto et al. aims to train policies that are robust to the sim-to-real gap by interpreting dynamics model errors of the simulator as an additional noise and jointly training an adversary and a protagonist \cite{pinto2017robust}. The adversary's goal is to learn adversarial actions that are added to the protagonist's actions at each time step as additional noise. The adversary's reward at each time-step is the negative reward of the protagonist, thus, encouraging the adversary to apply harmful noise. By training the protagonist in the presence of such an adversary, the protagonist is required to learn a policy that is robust to harmful perturbations, making it more robust to model inaccuracies of the simulator and, thus, to the sim-to-real-gap. However, in contrast to our method, such approach is in general not able to steer the agent to far away states and thus does not tackle initial state distribution mismatch. Our method is different to RARL as our method does not train the adversary to add noise to the protagonist's actions, but steer the agent to difficult states from which the protagonist takes over to solve the task without any additionally generated interference by the adversary.

\textbf{Curriculum learning.}
Curriculum learning methods learn to solve tasks in a variety of contexts by presenting context-based tasks to the learning agent in an order of increasing complexity instead of presenting tasks randomly \cite{bengio2009curriculum}. Held et al. train a generative adversarial net (GAN)  to produce increasingly difficult contexts \cite{held2017automatic}. Florensa et al. learn a reverse curriculum for goal-oriented tasks, in which a goal state is provided that needs to be reached from any initial state in the environment \cite{florensa2017reverse}. The idea of this method is to  maintain a set of initial positions that is initialized to the provided set of goal states. New candidates for initial states are generated by random walks beginning from states of the current set. These candidates are added to the set if they have the desired difficulty according to a simple heuristic or dropped if deemed too easy. This approach makes the assumptions that the environment's goal-state is known during training and that the environment can be reset to arbitrary states in the environment, which our approach does not. The POET algorithm uses evolution strategies to generate a range of related environments to learn policies that can transfer from one environment to another \cite{wang2019paired}.
Sukhbaatar et al. train an adversary to create an automatic curriculum for the protagonist by interleaving standard RL training with adversarial self-play episodes that provide an intrinsic motivation to explore the environment \cite{sukhbaatar2017intrinsic}. This context-based approach splits the training process into two different types of episodes. In self-play episodes, an adversarial policy controls the agent and interacts with the environment until it selects a stop-action. At this point in the episode, the protagonist takes over and is tasked with either reversing the adversary's trajectory or to replay it, the desired state being provided to the policy as context. During self-play episodes, no extrinsic reward is provided to the agents: the adversary is rewarded positively for each time step that the protagonist requires to reach the desired state and negatively for each own action. This reward structure encourages the adversary to generate increasingly difficult self-play scenarios for the protagonist. The protagonist's reward in self-play episodes is the negative number of steps required to reach the desired state. The second type of episodes simply consist of the target task that the protagonist is expected to learn. The context in such episodes is set to zero and an additional flag is set to inform the policy of current type of episode. We attempted to test the self-play approach in our maze environment. However, we were unable to achieve better results than standard SAC in the continuous maze environment despite best efforts. We hypothesize that the approach does not scale well to continuous state and action spaces. These findings are in line with the observations made by \cite{florensa2017reverse}.

\section{ADVERSARIAL REINFORCEMENT LEARNING}
\label{sec:arl}

We consider an irreducible MDP with state space $\mathcal{S}$, action space $\mathcal{A}$, system dynamics $\mathcal{P}(\mathbf{s}'|\mathbf{s}, \mathbf{a})$, initial state distribution $\mu(\mathbf{s})$, reward function $r(\mathbf{s}, \mathbf{a}, \mathbf{s}')$, discount factor $\gamma$ and a finite horizon $H_p$. The goal of RL methods is to find a parameterized policy $\pi_{\mathbf{\theta}}(\mathbf{a}|\mathbf{s})$ that maximizes the accumulated discounted reward 
when sampling the state at time step zero from the initial distribution $\mu(\mathbf{s})$ and afterwards following the policy $\pi_\theta$ and system dynamics $\mathcal{P}$, i.e., 
\begin{equation}
    \label{eq:rl_objective}
    J(\pi) = \mathbb{E}_{\mu, \pi_\theta, \mathcal{P}} \left[\sum_{t=0}^{H_p-1} \gamma^{t} r(\mathbf{s}_t, \mathbf{a}_t, \mathbf{s'}_t) \right].
\end{equation}

However, we assume that the initial state distribution $\mu(\mathbf{s})$ is unknown and resetting the robot to it is infeasible. Instead, we assume that the environment can only be reset according to a reset distribution $\mu_R(\mathbf{s})$ which is typically less diverse and less challenging. For example, we want to solve the robot disentangling task for a wide range of entanglements but are only able to reset the robot (and its attached object) to a small number of manually specified positions.

The main idea of our approach is to reduce the distribution mismatch (covariate shift) between the reset distribution and the unknown initial distribution by steering the agent into regions in which the current policy performs poorly. To steer the agent into these regions, we train an adversarial agent that chooses the protagonist's initial position by interacting with the environment. The adversary's task is to find areas in the environment's state space that are challenging for the protagonist. The adversary is rewarded for moving the agent to situations that are expected to yield low returns for the protagonist under its current policy. At the beginning of each episode, the agent is controlled by the adversarial policy for a fixed number of time steps after which the protagonist takes over again. The adversary's and protagonist's policies are trained jointly to allow the adversary to adapt to the changing protagonist policy. 

Our learning framework consists of an MDP environment $E$ of horizon $H_P$ and two policies, the \textit{protagonist} $\pi_P$ and the \textit{adversary} $\pi_A$, which both interact with the same environment. The ARL framework aims to learn a protagonist's policy $\pi_P$ that performs well in as many areas of the environment as possible. We extend the original RL framework, to encourage directed exploration of difficult areas. At the beginning of each training episode, the agent acts for $H_A$ time steps under policy $\pi_A$ and subsequently follow the protagonist policy $\pi_P$ for $H_P$ time steps. We can formulate the protagonist's objective function in the following way:

\begin{equation}
    \label{eq:protagonist_objective}
    J(\pi_P) = \mathbb{E}_{\mu_P, \pi_P, \mathcal{P}} \left[\sum_{t=0}^{H_{P}-1} \gamma^{t} r(\mathbf{s}_t, \mathbf{a}_t, \mathbf{s'}_t) \right],
\end{equation}

where the protagonist's initial state probability $\mu_{P}(\mathbf{s}_0)$ is produced by executing the adversary for $H_A$ steps starting from a state that is sampled from the reset distribution $\mu_R(\mathbf{s})$.  

The adversary's goal is to find states $\mathbf{s}$ in the environment for which the protagonist's expected return is low. The protagonist's state-value function $V^{\pi}$ estimates the expected return of following policy $\pi$ starting in state $\mathbf{s}$. Thus, we can define a reward function $r(\mathbf{s}, \mathbf{a}, \mathbf{s'})$ that evaluates the adversary's action through the protagonist's negative state-value of the reached state, that is, 

\begin{equation}
r_A(\mathbf{s}_t, \mathbf{a}_t, \mathbf{s'}_t;\pi_P) = -V_P( \mathbf{s'}), 
\end{equation}

where $V_P$ is the current protagonist's estimated state-value function. It is to note that we do not only provide reward based on the adversary's final state but also for immediate steps in order to provide a more shaped reward. With this choice of reward function, the adversary is encouraged to explore the environment to find states in which the protagonist's policy is expected to perform poorly. We can formulate the adversary's objective as

\begin{equation}
    J(\pi_A) = \mathbb{E}_{\mu, \pi_A, \mathcal{P}} \left[\sum_{t=0}^{H_{A}-1} \gamma^{t} r_A(\mathbf{s}_t, \mathbf{a}_t, \mathbf{s'}_t; \pi_P) \right]. 
\end{equation}

We jointly train both agents by alternating between improving the adversary's policy and the protagonist's policy. Figure~\ref{fig:arl_vis} presents a high-level visualization of the ARL learning framework. Our implementation is based on the soft actor-critic~\cite{haarnoja2018soft2} RL algorithm which is off-policy and, thus, we describe our implementation for off-policy reinforcement learning. It is, however, also possible to use the ARL framework in combination with on-policy RL methods. Training begins by updating the adversary for $K_A$ episodes. At the start of each episode, the environment is reset to a state sampled from $\mu_{R}$. Next, at each timestep $t$, the adversary selects an action using its policy $\mathbf{a}_t^{(A)} \sim \pi_{A}$ and observes the next state $\mathbf{s}_{t+1}$. The adversary's reward is computed using the current approximation of the protagonist's state-value function, i.e. $r_t^{(A)} = -V_{P}(\mathbf{s}')$. The tuple $(\mathbf{s}_{t}, \mathbf{a}_{t}^{(A)}, r_{t}^{(A)}, \mathbf{s}_{t+1})$ is added to the adversary's replay buffer, from which a minibatch of transitions is sampled to update the adversary's policy. After $H_A$ steps, the protagonist begins to interact with the environment. Note that the environment is not reset before the protagonist begins its interactions, thus the protagonist's initial position is $\mathbf{s}_{H_A}$. The protagonist collects tuples $(\mathbf{s}_{t}, \mathbf{a}_{t}^{(P)}, r_{t}^{(P)}, \mathbf{s}_{t+1}) $ for $H_P$ time steps. The tuples are added to the protagonist's replay buffer $D_A$. However, the protagonist's policy is not updated during these rollouts. Once the protagonist has completed its interaction with the environment, the environment is reset. This process is repeated for $K_A$ episodes in which the adversary's policy is updated at each time step, while the protagonist simply collects experiences to store in its replay buffer $D_P$. After $K_A$ rollouts, it is the protagonist's turn to be improved. At the beginning of each rollout, the environment is reset. Next, the adversary interacts with the environment for $H_A$ time steps to generate the protagonist's starting position, while storing the experienced transitions in its replay buffer $D_A$. After the adversary has completed $H_A$ actions, it is the protagonist's turn again. For $H_P$ time steps {t}, the protagonist samples action $\mathbf{a}_{t}^{(P)} \sim \pi_{P}(\mathbf{s}_t)$, observes the new state $\mathbf{s}_{t+1}$ and receives the reward $r_{t}^{(P)}$. Next, the tuple $(\mathbf{s}_{t}, \mathbf{a}_{t}^{(P)}, r_{t}^{(P)}, \mathbf{s}_{t+1}) $ is added to the replay buffer $D_P$ and the protagonist's policy $\pi_P$ and state-value function approximation $V_P$ are updated using a minibatch of samples drawn from $D_P$. This sequence of alternating between training the adversary and training the protagonist is repeated for a fixed number of iterations $N$. It should be noted that the ARL framework introduces three additional hyperparameters: $K_A$, $K_P$ and $H_A$. Furthermore, it is possible to use any off-policy RL method that approximates the protagonist's state value function $V_P$ in this learning framework. Algorithm~\ref{alg:off_policy_ARL} shows the pseudocode of our proposed method.

\begin{algorithm}[h]

\caption{Off-policy adversarial reinforcement learning}
\SetKwData{X}{x}
\SetKwData{Dp}{$D_{P}$}
\SetKwData{Da}{$D_{A}$}
\SetKwData{Pia}{$\pi_{A}$}
\SetKwData{Pip}{$\pi_{P}$}
\SetKwFunction{Train}{train}
\SetKwFunction{Step}{step}
\SetKwFunction{Reset}{reset}
\SetKwInOut{Input}{Input}

\Input{Arbitrary initial policies \Pia and \Pip with replay buffers \Da and \Dp, Environment $E$, }

\For{$i \leftarrow 0$ \KwTo $N$}{
    \For{$j \leftarrow 0$ \KwTo $K_A$}{
        \Reset{$E$} \;
        \For{$t \leftarrow 0$ \KwTo $H_A$}{    
            $(\mathbf{s}_{t}, \mathbf{a}_{t}^{(A)}, r_{t}^{(A)}, \mathbf{s}_{t+1}) \leftarrow$ \Step{$E$, \Pia})\;
            \Da $\leftarrow$ \Da $\cup$ $(\mathbf{s}_{t}, \mathbf{a}_{t}^{(A)}, r_{t}^{(A)}, \mathbf{s}_{t+1})$\;
            \Pia $\leftarrow$ \Train{\Pia, \Da} \;
        }
        \For{$t \leftarrow 0$ \KwTo $H_P$}{    
            $(\mathbf{s}_{t}, \mathbf{a}_{t}^{(P)}, r_{t}^{(P)}, \mathbf{s}_{t+1}) \leftarrow$ \Step{$E$, \Pip})\;
            \Dp $\leftarrow$ \Dp $\cup$ $(\mathbf{s}_{t}, \mathbf{a}_{t}^{(P)}, r_{t}^{(P)}, \mathbf{s}_{t+1})$\;
        }
    }
    
    \For{$j \leftarrow 0$ \KwTo $K_P$}{
        \Reset{$E$} \;
        \For{$t \leftarrow 0$ \KwTo $H_A$}{    
            $(\mathbf{s}_{t}, \mathbf{a}_{t}^{(A)}, r_{t}^{(A)}, \mathbf{s}_{t+1}) \leftarrow$ \Step{$E$, \Pia})\;
            \Da $\leftarrow$ \Da $\cup$ $(\mathbf{s}_{t}, \mathbf{s}_{t}^{(A)}, r_{t}^{(A)}, \mathbf{s}_{t+1})$\;
        }
        \For{$t \leftarrow 0$ \KwTo $H_P$}{    
            $(\mathbf{s}_{t}, \mathbf{a}_{t}^{(P)}, r_{t}^{(P)}, \mathbf{s}_{t+1}) \leftarrow$ \Step{$E$, \Pip})\;
            \Dp $\leftarrow$ \Dp $\cup$ $(\mathbf{s}_{t}, \mathbf{a}_{t}^{(P)}, r_{t}^{(P)}, \mathbf{s}_{t+1})$\;
            \Pip $\leftarrow$ \Train{\Pip, \Dp} \;
        }
    }
}

\label{alg:off_policy_ARL}
\end{algorithm}  

\section{EXPERIMENTAL RESULTS}
\label{sec:experiments}
\subsection{Environments}

\subsubsection{Object disentangling}
In the first environment, which we call the \textit{Robot-arm disentangling environment} (RADE), the agent controls a seven degree-of-freedom KUKA LBR iiwa R820 robotic arm which is equipped with a SAKE EzGripper end-effector, holding an "S"-shaped object which is entangled with a second, "O"-shaped object. Figure ~\ref{fig:hardware_setup} shows the experimental hardware setup. The MDP's state space is represented as a 7-dimensional vector of the current robot joint positions. The agent's goal is to disentangle the two objects through direct joint actions, i.e. providing joint deltas for all seven joints at each time step, in a fixed number of steps without colliding. The task solved if the euclidean distance $d$ between the objects' centers exceeds a threshold of 0.5 meters. The agent receives a positive reward for successfully solving the disentangling task. Collisions are penalized with a negative reward that depends on the current time step, where early collisions are punished more severely than later ones. Otherwise, a small action penalty is given to the agent to encourage smaller steps. Thus, the reward function for the disentangling task is formalized as,

\begin{equation}
    r(\mathbf{s},\mathbf{a},\mathbf{s'}) = 
    \begin{cases} 
    -\frac{1-\gamma^{H-t+1}}{1-\gamma},              & \text{if collision detected}, \\    
    1, & \text{if } d \geq 0.5, \\
    -\lVert \mathbf{a}\rVert_{2},              & \text{otherwise}. \\
    \end{cases}
\end{equation}

\begin{figure}
\vspace{1em}
\setlength\tabcolsep{2pt}
\centering
\begin{tabular}{cccc}
 \includegraphics[width=0.22\linewidth]{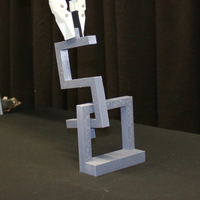} &
 \includegraphics[width=0.22\linewidth]{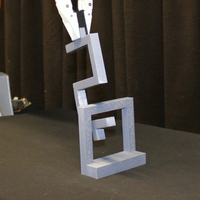} &
 \includegraphics[width=0.22\linewidth]{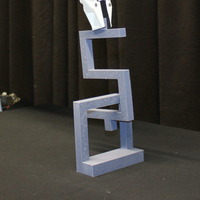} &
 \includegraphics[width=0.22\linewidth]{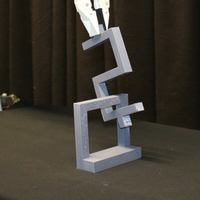} \\
 \includegraphics[width=0.22\linewidth]{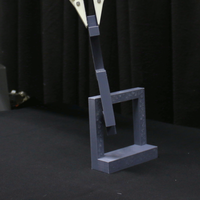} &
 \includegraphics[width=0.22\linewidth]{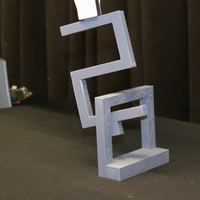} &
 \includegraphics[width=0.22\linewidth]{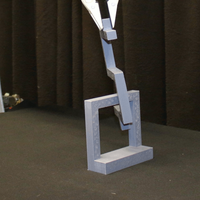} &
 \includegraphics[width=0.22\linewidth]{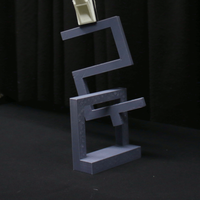}
\end{tabular}
\caption[The real robot in the starting positions form the training and test sets]{The top row of images shows the 4 initial positions from the training set, the bottom row the entanglements from the test set. Note that for both sets, the S-shape is positioned in the two top corners of the O-shape, once from each side. The test set consists of slight variations of the positions from the training set.}
\label{fig:start_positions}
\end{figure}

We evaluated the learned protagonist polices in two distinct ways to measure and compare their respective performances. The first method evaluates policies in simulation on a test set of initial positions. This test set consists of four different joint configurations in which the objects are entangled. These configurations have not been used as initial state for training. To measure how well an agent can generalize to new situations, we evaluate the protagonist's performance over 100 episodes while sampling the initial position uniformly from this test set. Figure~\ref{fig:start_positions} shows the possible initial position from both the training set and the test set.
Additionally, we tested the learned policy's performance by evaluating it in the real world environment. The initial states were sampled from the same distribution as during training, i.e. $\mu(\mathbf{s}) = \mu_R(\mathbf{s})$.

\subsubsection{Continuous maze}
\begin{figure}
      \centering
      \includegraphics[width=0.95\linewidth]{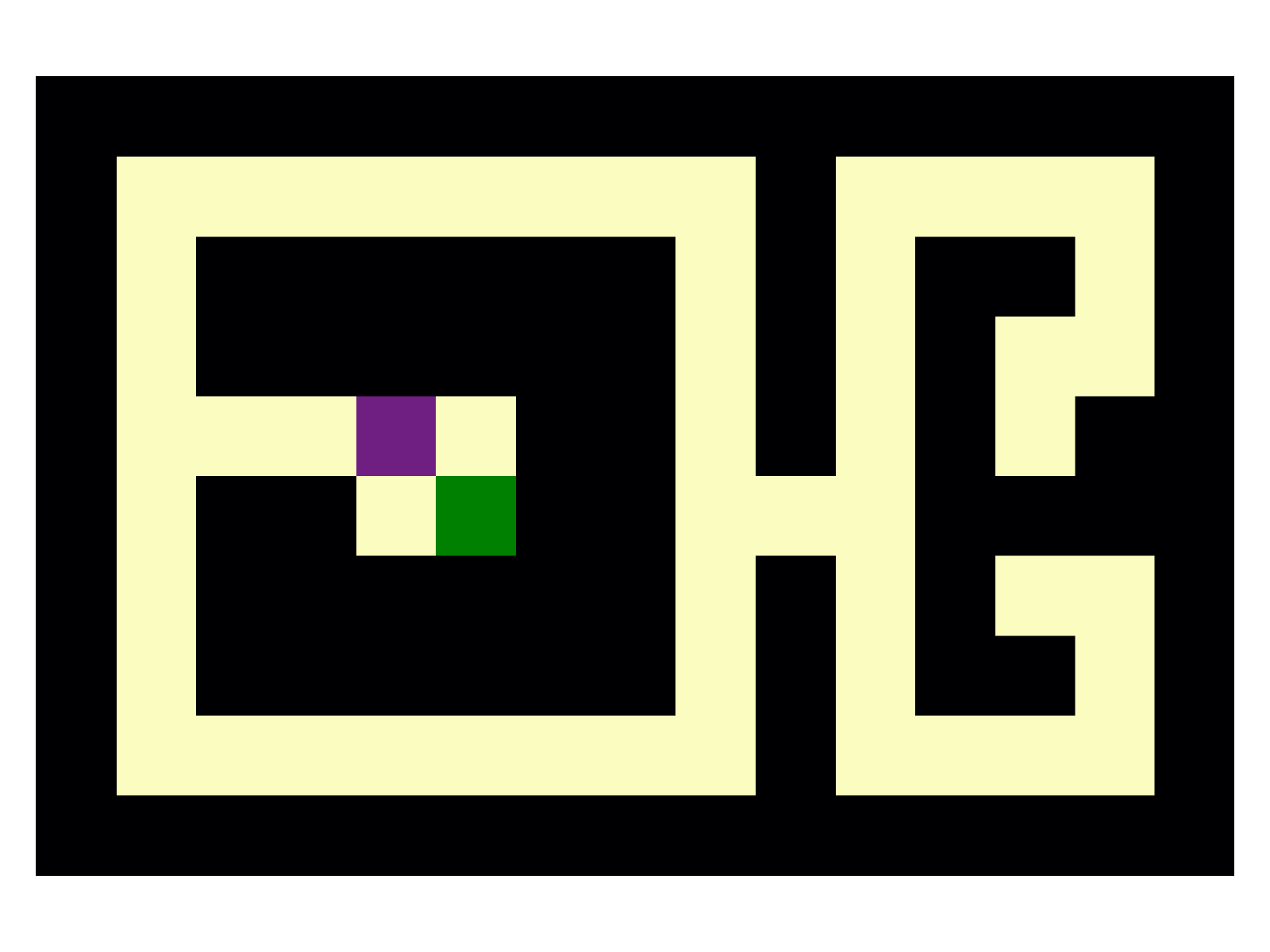}
    \caption[] {Illustration of the continuous maze environment. The agent's task is to navigate a point mass to the green goal square. Actions are continuous position deltas, observations are continuous 2D coordinates. In real-world environments, creating meaningful training scenario distributions requires domain knowledge. Instead, generating initial states close to the tasks goal state requires little domain knowledge and human labor. Thus, during training, the agent's start position is sampled from the reset distribution (purple). The adversary then moves to another state which becomes the initial state for the protagonist. The task's true initial state distribution, the uniform distribution over all states in the maze (light yellow), is only used for evaluation.}
    \label{fig:maze_mini}
\end{figure}

Additionally, we evaluated our approach in a continuous maze environment, in which the agent must navigate a complex maze to reach a goal position. While this toy example does not require DRL methods to be solved and could easily be trained uniformly on the complete state space in simulation, we chose to examine the ARL method on this task to easily visualize how the addition of an adversary affects the protagonist's performance over time. Furthermore, it shows that our method is applicable to a variety of tasks other than object disentangling. In the maze environment, the agent observes the current state as continuous 2D coordinates in the maze. The available actions to the agent are continuous velocities which are capped at a maximum of 1 square per time step. The agent receives a positive reward if the goal square is reached or a small negative movement penalty otherwise. Actions that would result in collisions with the maze walls ignore the component of the action that causes the collision. The agent's task in this environment is to reach the goal position from any point in the maze within 100 time steps. During training, only a limited set of training scenarios is available, as the initial position of the agent is sampled from a single square close to the goal. The maze environment is illustrated in Figure \ref{fig:maze_mini}.
\subsubsection{Research questions}
We designed our experiments in both environments to answer the following research questions:

\begin{enumerate}
    \item Does ARL generalize better than standard RL?
    \item How robust is ARL w.r.t. its hyperparameters?
\end{enumerate}

\subsection{Implementation details}

In order to facilitate experiments, we created a simulated version of the disentangling environment. The existing robot simulator and controller \textit{Simulation Lab} (SL) \cite{schaal2009sl} was used, both, for simulation and to control the real robots. Furthermore, we implemented our environments as extensions to the existing \textit{OpenAI gym} environments \cite{brockman2016openai}. Using this standardized environment interface allowed us to easily use readily available implementations of state-of-the-art RL methods like Soft Actor-Critic (SAC) from the \textit{OpenAI baselines}  \cite{dhariwal2017openai} and \textit{stable-baselines} \cite{hill2018stable} projects for our experiments.

\subsection{Results}

To answer the previously stated research questions, we carried out a small sensitivity study, comparing various hyperparameter settings. Our main focus in these experiments was to investigate how the number of training episodes per iteration, $K_A$ and $K_P$, and the adversary's horizon, $H_A$, affect the learning process. To analyze the effects of these hyperparameters, we trained three different variants of adversarial SAC with varying choices of $K_A$ and $K_P$. Hereby, we chose to set $K_A = K_P$ to allow an equal number of training episodes for both protagonist and adversary:

\begin{itemize}
 \item \textbf{ASAC10}: Adversarial SAC with $K_A = K_P = 10$
 \item \textbf{ASAC100}: Adversarial SAC with $K_A = K_P = 100$
 \item \textbf{ASAC1000}: Adversarial SAC with $K_A = K_P = 1000$
\end{itemize}

Note that all policies were trained for the same number of episodes. Thus, the number of training iterations $N$ depends on the choice of $K_A$ and $K_P$, i.e. to train ASAC10 for ten thousand episodes, one thousand iterations were required, while training ASAC1000 for the same amount of episodes only 10 training iterations are required.

\subsubsection{Disentangling results}

In the disentangling environment, we tested all three variants of ASAC using three different horizons for the adversary and compared them against a baseline of standard SAC. In the following, we will present and discuss the results of this sensitivity study. We used feedforward neural network policies with 2 hidden layers of 64 units for all disentangling experiments. All SAC hyperparameters were set equally, performing 5 optimization steps for every step in the environment with a learning rate $\alpha$ of 0.0003. Each policy was trained for ten thousand episodes. 

\begin{table}
\centering
\caption{Mean return and success rate ($\pm$ standard error), over 10 evaluation runs, for SAC and ASAC on the training and test sets. Maximum values are in bold font. ASAC outperforms SAC on the test set.}
\resizebox{0.95\columnwidth}{!}{
\begin{tabular}{l|c c|c c}
\firsthline
 \multicolumn{1}{c}{Method} & \multicolumn{2}{c}{Training} & \multicolumn{2}{c}{Test set} \\
 & return &  success rate (\%) & return & success rate (\%) \\
\hline
SAC & $0.44 \pm 0.29$ & $90.5 \pm 4.84$ & $-3.90 \pm 0.39$ & $17.0 \pm 6.63$ \\
\hline
ASAC10 ($H_A=20$) & $0.20 \pm 0.51$ & $86.4 \pm 8.65$ & $-3.71 \pm 0.60$ & $20.20 \pm 10.11$ \\
ASAC100 ($H_A=20$)& $0.07 \pm 0.43$ & $84.2 \pm 7.31$ & $-2.60 \pm 0.72$ & $39.00 \pm 12.27$ \\
ASAC1000 ($H_A=20$)& $-0.42 \pm 0.63$ & $76.0 \pm 10.67$ & $-3.83 \pm 0.45$ & $18.2 \pm 7.64$ \\
\hline
ASAC10 ($H_A=5$)& $-0.18 \pm 0.35$ & $80.0 \pm 6.01$ & $-3.57 \pm 0.41$ & $22.6 \pm 6.96$ \\
ASAC100 ($H_A=5$)& $-0.51 \pm 0.49$ & $74.34 \pm 8.28$ & $-3.03 \pm 0.43$ & $31.78 \pm 7.26$ \\
ASAC1000 ($H_A=5$)& $-0.26 \pm 0.25$ & $78.6 \pm 4.25$ & $-2.74 \pm 0.37$ & $36.7 \pm 6.21$ \\
\hline
ASAC10 ($H_A=1$)& $0.30 \pm 0.29$ & $88.1 \pm 4.99$ & $\mathbf{-2.56 \pm 0.31}$ & $\mathbf{39.7 \pm 5.27}$ \\
ASAC100 ($H_A=1$)& $\mathbf{0.70 \pm 0.05}$ & $\mathbf{94.88 \pm 0.83}$ & $-3.01 \pm 0.28$ & $32.13 \pm 4.78$ \\
ASAC1000 ($H_A=1$)& $0.00 \pm 0.33$ & $83.1 \pm 5.60$ & $-2.81 \pm 0.36$ & $35.0 \pm 6.39$ \\

\end{tabular}}

\label{tbl:disentangling_results_sim}

\end{table}

Table~\ref{tbl:disentangling_results_sim} shows the performance of the learned protagonist policies on both the training set and the test set in simulation. The left side of the table shows the mean scores and percentages of solving the task of the final 100 episodes of training. Despite the presence of an adversary, all variants of ASAC achieve a similar performance to SAC on the training set. We can observe that ARL leads to policies that are able to solve the majority of examples presented to it. The performance of most ASAC methods is lower than that of SAC since SAC policies were trained on the ideal initial state distribution. We evaluated the final policies of 10 runs on the test set (25 trials per test scenario) in simulation. We can observe that all ASAC policies achieve better performance than SAC. ASAC using a 1-step adversary can solve the presented test scenario twice as often as SAC. We hypothesize that the slightly better performance of ASAC with a 1-step adversary, compared to those with larger horizon is caused by the similarity of the training and test scenarios, thus only a single action of the adversary generates a new meaningful scenario. While larger horizons enable the adversary to potentially find a greater variety of difficult scenarios, they also require the adversary to explore areas of the environment that are trivial to solve for the protagonist, e.g. already disentangled positions.

\begin{table}
\centering
\caption[]{Mean return and success rate ($\pm$ standard error), over 5 evaluation runs on the real robot, for SAC and single step ASAC10 on the training and test sets. ASAC outperforms SAC on the training set and the test set.}
\resizebox{0.95\columnwidth}{!}{
\begin{tabular}{l|c c|c c}
\firsthline
 \multicolumn{1}{c}{Method} & \multicolumn{2}{c}{Training} & \multicolumn{2}{c}{Test set} \\
 & return & success rate (\%) & return & success rate (\%) \\
\hline 
SAC & $-0.44 \pm 0.55$ & $80.0 \pm 9.35$ & $-4.31 \pm 0.36$ & $10.0 \pm 6.12$ \\
\hline
ASAC10 & $ \mathbf{0.41 \pm 0.59}$ & $\mathbf{90.0 \pm 10.00}$ & $\mathbf{-2.44 \pm 1.02}$ & $\mathbf{41.67 \pm 17.28}$ \\
\end{tabular}
}
\label{tbl:results_real_robot}
\end{table}

We chose the method that showed the best performance on the test set in simulation, namely ASAC10 with $H_A = 1$, to be evaluated and compared against SAC in the real-world environment on both the train and test set. We evaluated five learned policies of each method for three trials per scenario, totalling 120 rollouts on the real robot. Table \ref{tbl:results_real_robot} shows the mean score and success rate of the policies. We can observe that ASAC not only solves the test scenarios 4 times as often, but also outperforms SAC on the training set. We believe that, due to the extended training set generated by the adversary, the protagonist was able to learn a more general policy that is more robust to the sim-to-real gap than SAC. Furthermore, the presence of an adversary improved the learned policy's performance on the test set in all disentangling experiments, indicating that ARL helps alleviate the problem of overfitting in DRL. Although simpler methods to generate extended training sets exist, e.g. random shaking at the beginning of each episode, such methods generally require domain knowledge about the given task and do not steer training to areas that require improvement. The ARL framework required no domain knowledge and actively forces the current policy's weak points to be improved.

\subsubsection{Maze results}

In the following, we present the findings of the evaluation of the ARL framework in the maze environment. We evaluated the same three variants of ASAC and compared their performance against standard SAC and an additional baseline using a random adversary (RA). In all experiments, we set $H_A$ to 100 time steps, allowing the adversary to reach any point in the maze regardless of its start position. We evaluate the protagonist's learnt policy by testing its performance from uniformly distributed starting positions over the complete state space.

\begin{table}
\centering
\caption{Mean return and success rate ($\pm$ standard error), over 5 evaluation runs, for SAC, ASAC and RA. Maximum values are in bold font. ASAC outperforms SAC and RA.}
\resizebox{0.95\columnwidth}{!}{
\begin{tabular}{l|c c}
\firsthline
Method & return &  success rate (\%)  \\
\hline
SAC & $-3.47 \pm 0.16$ & $27.44 \pm 3.23$ \\
Random Adversary & $-3.43 \pm 0.19$ & $28.08 \pm 3.67$ \\
\hline
ASAC10 & $-2.21 \pm 0.56$ & $55.60 \pm 12.19$ \\
ASAC100 & $-2.44 \pm 0.59$ & $51.33 \pm 12.90$ \\
ASAC1000 & $\mathbf{-2.08 \pm 0.63}$ & $\mathbf{59.21 \pm 13.97}$ \\
\end{tabular}}

\label{tbl:results_maze}

\end{table}
\begin{figure}
\centering
\includegraphics[width=0.99\linewidth]{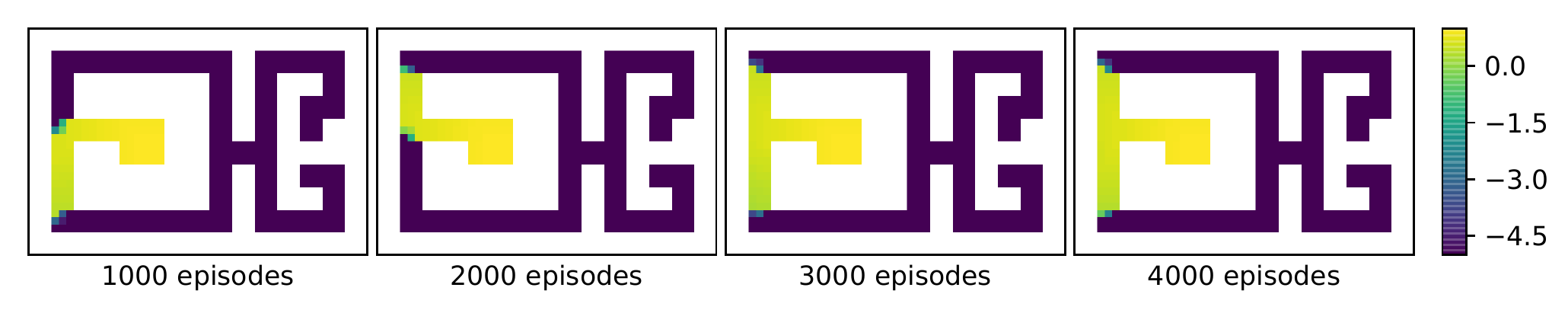}
\includegraphics[width=0.99\linewidth]{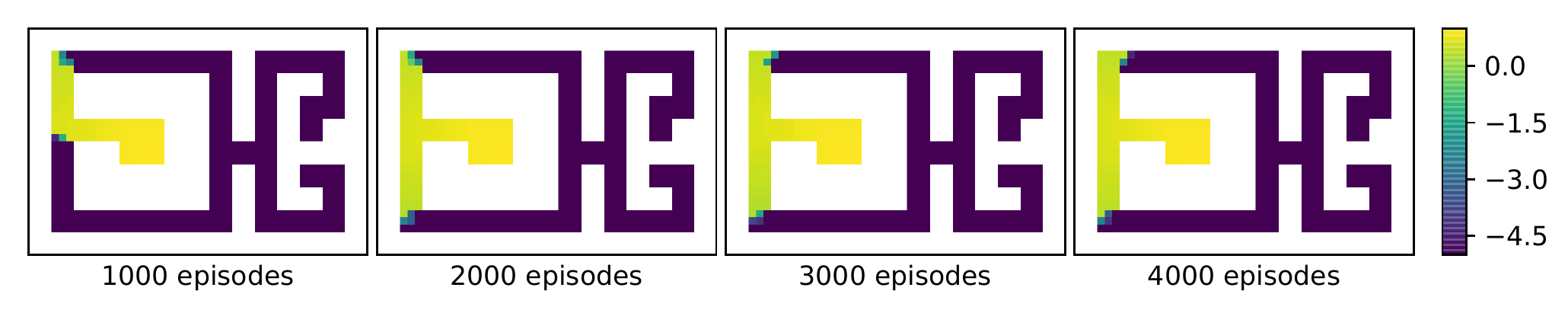}
\includegraphics[width=0.99\linewidth]{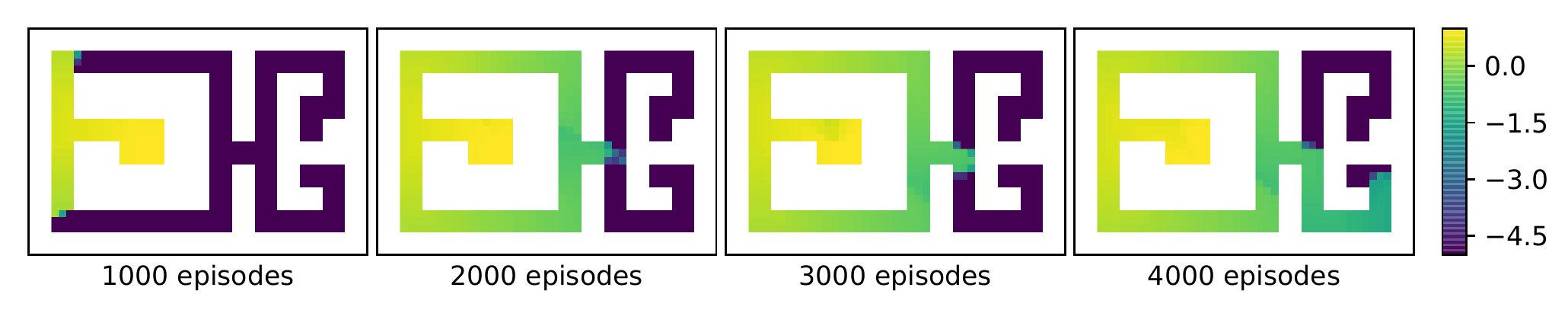}

\caption[]{Example progression of mean return after 1000, 2000, 3000 and 4000 training episodes depending on the agents initial position. Top row shows the random adversary baseline, middle shows SAC and the bottom row shows ASAC10. Guided by the adversarial policy, ASAC continuously progresses to explore distant regions of the maze. SAC and RA only explore close to the reset square.}
\label{fig:progression_maze}
\end{figure}
Table \ref{tbl:results_maze} shows the mean episode return and success rate on the complete maze of ASAC, SAC and RA. We can observe that all ASAC methods clearly outperform SAC and RA, being able to solve the maze task from twice as many different start positions. While SAC only explores the maze close to the reset state, the random walk adversary is not able to extend the training set of scenarios in a way that aids the protagonist's learning process. Figure \ref{fig:progression_maze} shows heatmaps of the mean return of the protagonist the training process of ASAC, SAC and RA. While all methods are able to learn to solve task from the left side of the maze, only ASAC learns a policy that is able to solve the maze from almost everywhere in the maze. These results indicate that the adversary successfully pushes the protagonist to difficult situations resulting in a more general protagonist policy. 
\section{CONCLUSION}
\label{sec:conclusion}
This paper introduced a novel adversarial learning framework, ARL, that trains agents to learn more general policies by steering the agent to difficult regions of the environment through adversarial policy. The approach implicitly generates new training scenarios for the protagonist, making the protagonist less likely to overfit the predefined set of training scenarios and more robust to the sim-to-real gap. We evaluated ARL on the robot disentangling task and the continuous maze. Our experiments included a sensitivity study to investigate the effects of changing hyperparameters, namely the adversary's horizon $H_A$ and the number of training episodes per iteration $K_A$ and $K_P$ for each policy. The results of the sensitivity study show that the performance of ARL is robust to the choice of $K_A$ and $K_P$ as all tested choices led to significantly improved performance on the test scenarios. The adversary's horizon impacts the learning process as adversaries with too small horizons $H_K$ lack the ability to reach all parts of the environment's state space. However, even a single step adversary can improve the protagonist's ability to generalize. \\
Our results provide several natural starting points for potential future work. First, investigating how ARL performs using other base algorithms instead of SAC remains a question to be answered. Additionally, we plan on evaluating our method one additional tasks to examine its overall generality and to further investigate the effects of hyperparameter choices. Most notably, we intend to examine the effects of different ratios between adversary and protagonist training episodes on performance and learning speed. As our previous experiments used low-dimensional state representations such as robot joint positions, we plan to test our approach using high dimensional visual inputs, i.e. camera images, to learn more general policies w.r.t. object shapes. Another follow-up would be to incorporate a shared critic that evaluates the actions of both the protagonist and the adversary to further improve learning speed and data-efficiency. Finally, we plan to evaluate whether our approach is able to increase robustness against adversarial attacks.


\section*{ACKNOWLEDGMENT}
Calculations for this research were conducted on the Lichtenberg high performance computer of the TU Darmstadt.

\bibliographystyle{IEEEtran}
\bibliography{lit}


\end{document}